\documentclass{article}

\usepackage{PRIMEarxiv}

\usepackage[utf8]{inputenc} % allow utf-8 input
\usepackage[T1]{fontenc}    % use 8-bit T1 fonts
\usepackage{hyperref}       % hyperlinks
\usepackage{url}            % simple URL typesetting
\usepackage{booktabs}       % professional-quality tables
\usepackage{amsfonts}       % blackboard math symbols
\usepackage{nicefrac}       % compact symbols for 1/2, etc.
\usepackage{microtype}      % microtypography
\usepackage{lipsum}
\usepackage{fancyhdr}       % header
\usepackage{graphicx}       % graphics
\usepackage{natbib}
\usepackage{amsmath}
\graphicspath{{media/}}     % organize your images and other figures under media/ folder
\usepackage{pifont}

\newcommand{\cmark}{\ding{51}}
\newcommand{\xmark}{\ding{55}}

\title{VADSK: Video Anomaly Detection with Structured Keywords
}

\author{
  Thomas Foltz \\
  The Pennsylvania State University \\
  State College, PA\\
  \texttt{tjf5667@psu.edu} \\
}

\begin{document}
\maketitle

\begin{abstract}
This paper focuses on detecting anomalies in surveillance video using keywords by leveraging foundational models' feature representation generalization capabilities. We present a novel, lightweight pipeline for anomaly classification using keyword weights. Our pipeline employs a two-stage process: induction followed by deduction. In induction, descriptions are generated from normal and anomalous frames to identify and assign weights to relevant keywords. In deduction, inference frame descriptions are converted into keyword encodings using induction-derived weights for input into our neural network for anomaly classification.  We achieved comparable performance on the three benchmarks UCSD Ped2, Shanghai Tech, and CUHK Avenue, with ROC AUC scores of 0.865, 0.745, and 0.742, respectively. These results are achieved without temporal context, making such a system viable for real-time applications. Our model improves implementation setup, interpretability, and inference speed for surveillance devices on the edge, introducing a performance trade-off against other video anomaly detection systems. As the generalization capabilities of open-source foundational models improve, our model demonstrates that the exclusive use of text for feature representations is a promising direction for efficient real-time interpretable video anomaly detection.
\end{abstract}


\section{Introduction}
In our modern society, there is an increasing need for video anomaly detection to ensure public safety, prevent crime, and identify environmental hazards. As surveillance capabilities increase, especially in highly populated locations, there is a high demand for intelligent systems that can efficiently process large amounts of video data to identify anomalies \cite{yossef2023review}. The sheer volume of data has exceeded the human capacity for effective monitoring, which has led the machine learning research community to devote effort toward developing automated anomaly detection solutions.\\

Originally, video anomaly detection relied on separating feature extraction from the classification process. This proved to be limited when handling complex situations in the data; however, it proved beneficial as a foundation for emerging methods using deep learning techniques \cite{wu2024deep}. Many modern applications now leverage neural networks to learn representations from raw video data. This has been possible due to the emergence of benchmarks such as UCSD Ped2 \cite{mahadevan2010anomaly}, ShanghaiTech \cite{liu2018future}, and CUHK Avenue \cite{lu2013abnormal}. These datasets include diverse anomaly scenarios, labeling various events that exclude anomalies in the training data, and a blend of normal and anomalous events in the test data. This enables both one-class and binary classification tasks with supervised, semi-supervised, weakly supervised, and unsupervised learning depending on how the data are preprocessed \cite{yossef2023review}. Our work only uses the test dataset due to our supervised binary classification approach. Other commonly used benchmarks include UCF Crime \cite{sultani2018real} and XD-Violence \cite{wu2020not}, which focus on violent or criminal real-world events. These two datasets separate the types of anomalous events, allowing models to train specifically to identify a specific anomaly event type or to create a multiclass classifier.\\

Although the task has improved considerably over the past decade, some issues still hinder its real-world applicability. One-class classification tasks have difficulty capturing complexity and diversity between anomaly types. Since these implementations can only train on normal data, they can become sensitive to deviations, leading to high false positive rates \cite{wu2024deep}. Real-time methods have the issue of sacrificing accuracy for speed with oversimplified pipelines. In many cases, they still require video sequences as input, introducing latency that makes them unusable for applications requiring instant recognition of anomalies.\\

Another issue with deep learning techniques is that they are typically computationally expensive and lack interpretability \cite{wu2024deep}. The “black box” nature makes it difficult to interpret why certain examples are flagged as anomalous, which reduces the user's trust and ability to refine the models. Attempts have been made to create interpretable systems that use object detection, pose, or trajectories to justify the predictions. However, the problem is that most of these cannot be implemented in real-time and include complex data pipelines, rendering them unusable in an application. New methods employing Large Language Models (LLMs) address the interpretability issue by generating textual explanations for anomalies that users easily comprehend. These solutions are complex to implement, still struggle to provide concise explanations for decisions, and use extensive computing power.\\

\begin{figure}
\center\includegraphics[width=1.0\textwidth]{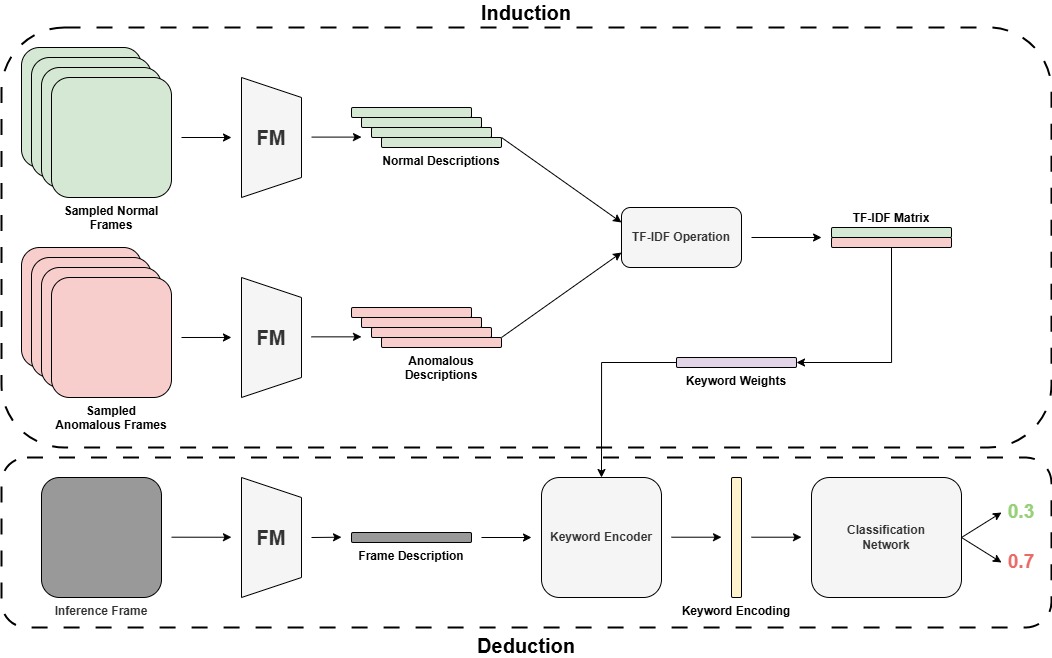}
\caption[Pipeline Overview. FM abbreviates the foundational models necessary for generating text descriptions from frame input. TF-IDF abbreviates the Term Frequency-Inverse Document Frequency score, which we use to weigh keywords.]{Pipeline Overview. FM abbreviates the foundational models necessary for generating text descriptions from frame input. TF-IDF abbreviates the Term Frequency-Inverse Document Frequency score, which we use to weigh keywords.}
\label{fig:pipeline_overview}
\end{figure}

This paper addresses these issues by employing a novel approach for interpretable video anomaly detection without reducing real-time performance or requiring large amounts of computing power. Our approach leverages foundational models' feature representation generalization capabilities to extract meaningful keywords from video. We employ a two-stage induction and deduction system as in a similar LLM-based solution \cite{yang2025follow}. 

As seen in Figure~\ref{fig:pipeline_overview}, we lay out the sequence of events between frame-level input and prediction. Before test time, the induction stage is performed to preprocess the data and learn feature representations in keyword weightings. Randomly sampled normal and anomalous frames are passed through our pre-trained foundational model to output corresponding descriptions of those frames. Next, we take the normal and anomalous descriptions and pass them through a TF-IDF operation, outputting a corresponding TF-IDF matrix of scores. This matrix stores values of the top $k$ relevant keywords based on how relevant the word is to its corpora of descriptions. With this information, we derive a vector of keyword weights based on how indicative those keywords are of an anomaly. In the deduction stage, we generate frame descriptions identically as in induction, but now the keyword weight vector is used to convert the frame description into a keyword encoding. For each keyword, if that keyword is present in the frame description, the corresponding element of the encoding is set to the weight found in the keyword weight vector. This keyword encoding can be directly passed into the classification network to output a probability prediction if that frame includes an anomaly. It is important to note that before test time, some data must be used to train the classification network. After the network has learned the decision boundary, the model is ready for inference using only the deduction stage of the pipeline.

This approach allows us to reduce the computational overhead of inference due to the minimal feature space and simple classification architecture, making it suitable for application on resource-constrained systems. By demonstrating the effectiveness of foundational models for interpretable video anomaly detection, this work creates new opportunities for developing transparent and trustworthy surveillance purposes.\\

The contributions of this paper are as follows:\\

\begin{enumerate}
\item Our findings show that a keyword-based approach can potentially identify video anomalies. This approach increases the user interpretability of decisions, improving transparency and trust in our model.

\item We introduce a lightweight, two-stage video anomaly detection pipeline based on induction followed by deduction. During the induction step, normal and anomalous frames are randomly sampled, from which sets of descriptions are generated. We calculate the term frequency-inverse document frequency (TF-IDF) score with these two sets to determine weights for the most relevant keywords. In the deduction step, descriptions are generated on the inference frames and encoded based on the keyword weights from the induction step. This encoding is then passed into a binary classification network for the final prediction.

\item Our methodology has achieved comparable performance to existing benchmarks while achieving near real-time inference, reduced model complexity, and decreased memory usage. This demonstrates the usability of our system for real-world applications that have constrained computing requirements and demand fast response times.
\end{enumerate}

\section{Related Works}
We reviewed current video anomaly detection approaches to identify areas for improvement in our system. We explored advancements in one-class, real-time, and interpretable video anomaly detection systems. Then, we thoroughly reviewed the emerging field that leverages large-language models' prediction capabilities for identifying video anomalies. Finally, we looked at some natural language processing techniques for classifying anomalies in text information. We did this to understand how to reapply older techniques with emerging technologies to improve current methodologies.

\subsection{One-class Classification Methods}

One of the original deep-learning-based detection techniques, one-class classification, detects when events occur outside of distribution. They exploit the vast amount of labeled normal data, unique to other systems that require labeled anomaly examples during training. One such method follows this idea by claiming that enhanced inliers and distorted outliers effectively decide anomalies. \citep{sabokrou2018adversarially} They employ dual reconstructor-discriminator architecture, where the reconstructor learns the concept of the normal class. This is done to reconstruct the normal samples correctly while distorting the anomalous samples that do not share the same concepts. The discriminator learns how to differentiate the two reconstructed image classes and make a prediction. Works such as HF2-VAD \citep{liu2021hybrid} focus on the flow reconstruction of objects in a frame. They predict the optical flow of previous frames and fuse that information with the current frame. A separate reconstructor module uses this information to predict the next frame. An anomaly is detected if the predicted future frame deviates more than expected, determined by a set threshold. Although the one-class classification technique has shown high performance, it tends to indicate false positives often in scenarios where new normal scenarios arise in the data. Additionally, it is difficult for people to understand the criteria determined outside of the normal distribution, making it less interpretable. 

\subsection{Real-time Methods}
Even though many video anomaly detection classifiers have effective discriminative capability, there have been efforts toward real-time classification to decrease the dependence on extended temporal contexts. One approach uses an end-to-end pipeline that learns features directly from raw video data to train their custom visual feature extractor rather than relying on commonly used pre-trained feature extractors that most modern methods use \citep{karim2024real}. Then, using k-nearest-neighbors (KNN) distances and uniform frame sampling, they train a lightweight classifier to predict anomalies in near-real time in a small decision window of approximately six seconds. Another approach for inference efficiency is introduced by MULDE \citep{micorek2024mulde}, by measuring how much feature vectors deviate from the normal distribution of frames, similar to how many one-class methods function. They train a classification model with different levels of injected noise into the training data to emulate anomalies. This can be done at the object or frame level. At test time, they used a Gaussian mixture model to combine these noise levels to identify different anomaly cases. They achieve near real-time inference due to their simple pipeline, including a feature extractor, feed-forward network, and Gaussian mixture model.

\subsection{Interpretable Methods}
Interpretable methods in video anomaly detection have gained interest due to the "black-box" nature of many classical methods. These new methods aim to increase transparency in the decision-making process, crucial for understanding and trusting the model outputs. One approach focuses on semantic embedding using scene graphs \citep{doshi2023towards}. It leverages relationships between objects in a scene to provide interpretability in the video. Text Empowered Video Anomaly Detection (TEVAD), increases accuracy and interpretability by fusing textual features with spatio-temporal information \cite{chen2023tevad}. This is achieved using frame captions to capture events' semantic meaning and visual features. Another work utilizes attribute-based representations, representing objects in a scene with velocity and pose information \cite{reiss2022attribute}. This information is then used to determine an anomaly score through density estimation. These interpretable methods provide valuable information on the decision-making process by incorporating semantic information, textual features, and attribute-based representations. However, these methods rely on long temporal chains of information to effectively identify anomalies and newer LLM-based methods for improving interpretability have emerged in recent years.

\subsection{Large Language Model (LLM) Methods}
To further advance interpretability while taking advantage of advances in machine learning, researchers have begun to employ large-language models (LLMs) to detect anomalies in video. One such paper uses video language models (VLMs) captioning capabilities to identify activities and objects in a scene that indicate normal and anomalous behavior \citep{yang2025follow}. They curate a list of rules that indicate which objects and activities are normal or anomalous. This is done by describing the normal behavior from the normal frames and with that information identifying the opposite anomalous behavior. With these rules, they match captions to these rules during inference and use an LLM to reason if anomaly conditions have been met. We employ a similar approach for identifying anomalies but introduce a simplified and explainable implementation for selecting anomaly keywords used in classification. Another practical approach was proposed by Holmes-VAD \citep{zhang2024holmes}, where they leverage LLM capabilities to explain why anomalies can occur in hour-long video sequences. They achieve this through a multi-modal LLM that encodes user text prompts, projected visual classes, and patch tokens that the temporal sampler has selected as noteworthy. While these methods leverage the powerful feature extraction capabilities of large language and multimodal models, these methods are quite expensive to deploy in practice due to the large compute requirements necessary, making them mostly unviable for applications on the edge.

\subsection{Natural Language Processing (NLP) Methods}
Natural Language Processing (NLP) methods have long been employed in classification tasks, extending into anomaly detection. The Term Frequency-Inverse Document Frequency (TF-IDF) score has proven to be a versatile and effective approach for this. TF-IDF is a statistical measure that evaluates the importance of a word within a document relative to a larger corpus. One work applies this scoring method to classify anomalies in process logs, treating each log entry as a distinct document within the broader corpus of all logs \citep{sandhu2022detecting}. This approach enables the identification of unusual terms or patterns that may signify anomalous behavior. Similarly, in the analysis of network switch logs, TF-IDF has been used as part of an approach by combining it with the log frequency and the log probabilities to calculate an abnormal score for different components of the logs, enhancing the overall precision of the identification of anomalies \cite{nam2024log}. TF-IDF is effective for this task because of its ability to emphasize words that deviate from the norm within a given context, making it well-suited for detecting anomalies.\\

\section{Approach}

\subsection{Induction}
In this first stage of the detection pipeline, it is necessary to identify keywords indicative of anomalies. There are three main steps during induction. We generate frame descriptions by sampling labeled normal and anomalous frames. They are then passed into a foundational model along with a prompt to generate the descriptions. These descriptions are separated into two corpora and passed into a term frequency-inverse document frequency vectorizer. With the output vector, we calculate the difference between the frequency of highlighted keywords and normalize it to achieve a final weighting vector for use in the deduction stage.

\subsubsection{Frame Description Generation}
We selected two foundational models that fulfill our requirements for frame description generation. These models must have multi-modal input that allows the user to pass in an image alongside a prompt since our raw data is split videos for processing at the frame level. Additionally, the model should be optimized for visual recognition and image reasoning to create meaningful frame descriptions. The model should be trained with minimal parameters or possess quantization capability to store weights and perform inference on edge devices. Finally, it is important to utilize open-source weights for the transparency and reproducibility of our implementation. Therefore, we selected the Llama-3.2-11B-Vision-Instruct model from Meta \citep{dubey2024llama} and the MiniCPM-V-2\_6-int4 model from openbmb \citep{hu2024minicpm}, both of which are available to the public on the HuggingFace platform \citep{huggingface}. \\

\begin{figure}
\center\includegraphics[width=1.0\textwidth]{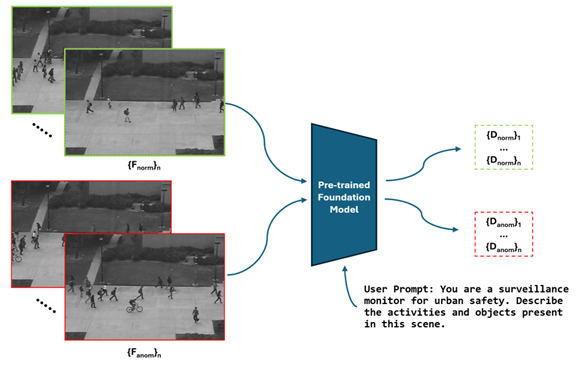}
\caption[Frame description generation. $F_{norm}$ represents labeled normal frames with their respective descriptions $D_{norm}$ and $F_{anom}$ represents labeled anomalous frames with their respective descriptions $D_{anom}$ generated by the pre-trained foundational model.]{Frame description generation. $F_{norm}$ represents labeled normal frames with their respective descriptions $D_{norm}$ and $F_{anom}$ represents labeled anomalous frames with their respective descriptions $D_{anom}$ generated by the pre-trained foundational model.}
\label{fig:frame_description_generation}
\end{figure}

With the selected models, we needed to generate captions that differentiate anomaly cases from normal scene behavior. To achieve this, we selected n frames randomly from the training data, which we know only includes normal frame samples. We then selected random frames known to be anomalies from the test set. This gives us an even sampling of normal and anomaly cases. As shown in Figure ~\ref{fig:frame_description_generation}, we pass those randomly sampled frames into our selected foundational model to generate the frame descriptions. This is passed in with the user prompt \textit{'You are a surveillance monitor for urban safety. Describe the activities and objects present in this scene.'} The first sentence in the prompt provides context to the foundational model of its task, and the second sentence explicitly asks for activities and objects so that we can extract meaningful keywords from the descriptions. We determined that sampling 20 normal and anomaly frames was sufficient to capture the most influential keywords in the corpora without over-fitting the data. \\

\subsubsection{Corpus Formation}
Once we have our two sets of generated frame descriptions, we concatenate the string descriptions in each set as depicted in equation~\ref{eq:corpus} to create a tuple representing the text corpora C. These corpora of two strings, one document for normal descriptions and one for anomalous descriptions, are passed into the TF-IDF vectorizer SKLearn library \citep{scikit-learn} for calculating the TF-IDF scores. \\

\begin{align}
\{D_{\text{norm}}\}_{i=1}^{n} & \text{: Set of } n \text{ description samples from normal frames.} \\
\{D_{\text{anom}}\}_{i=1}^{n} & \text{: Set of } n \text{ description samples from anomaly frames.} \\
D_{\text{norm\_joined}} &= D_{\text{norm}_1} D_{\text{norm}_2}\cdots D_{\text{norm}_n} \\
D_{\text{anom\_joined}} &= D_{\text{anom}_1} D_{\text{anom}_2}\cdots D_{\text{anom}_n} \\
C &= \{ D_{\text{norm\_joined}},  D_{\text{anom\_joined}} \}
\label{eq:corpus}
\end{align}

\subsubsection{Term Frequency-Inverse Document Frequency Score}
To identify how any term from the corpus relates to a corresponding document, we need to calculate the Term Frequency-Inverse Document Frequency score, a balance between measuring the frequency of a term occurring in a document and the amount that term shows up in any of the documents. \\

\begin{equation}
tfidf(t, d, C) = tf(t, d) \cdot idf(t, C)
\label{eq:tf-idf}
\end{equation}

\begin{equation}
tf(t, d) = \frac{f_{t,d}}{\sum_{t' \in d} f_{t',d}}
\label{eq:tf}
\end{equation}

\begin{equation}
idf(t, C) = \log \frac{N}{|\{d \in C: t \in d\}|}
\label{eq:idf}
\end{equation}

In equation~\ref{eq:tf-idf}, the overall TF-IDF score of term t in document d within the corpus C is calculated by multiplying each term’s term frequency score by the inverse-document frequency. In equation~\ref{eq:tf}, we determine the frequency of term t in document d by calculating the number of times t appears in d divided by the total number of terms in document d. Finally, in equation~\ref{eq:idf} we obtain the inverse document frequency of term t in corpus C, calculated with the logarithm of the total number of documents N in C divided by the number of documents containing t. After this TF-IDF vectorization operation on our corpus, we are left with two vectors: one for the scores of terms associated with the normal descriptions and one for the associated scores of the anomaly descriptions. \\

\subsubsection{Anomaly Keyword Weighting}
We derive a normalized difference vector by calculating the difference between the TF-IDF score vectors of the anomaly documents and the normal documents to identify terms indicative of anomalies. We normalize this difference vector to ensure that the magnitude of the differences does not skew the analysis and for stable classification training. The resulting vector highlights the terms more characteristic of the anomaly frame descriptions than the normal frame descriptions, providing insights into the keywords that distinguish anomalies.

\begin{equation}
w_{keywords}(t) = \frac{w_{diff}(t)}{||w_{diff}||} = \frac{w_{anom}(t) - w_{norm}(t)}{\sqrt{\sum_{t' \in T} (w_{anom}(t') - w_{norm}(t'))^2}} \quad \forall t \in T
\label{keyword_weighting}
\end{equation}

\subsection{Deduction}
In this second stage of the detection pipeline, we predict whether an anomaly was found frame by frame. The deduction stage has two main steps: creating a keyword encoding from the generated frame description and passing that encoding into a classification model for predicting the anomaly probability.

\subsubsection{Keyword Encoding}
First, we take a frame from a video we intend to infer. It is passed through the same pre-trained foundational model with identical user prompting as in induction. The difference is that instead of forming a corpus with a combination of frame descriptions as in Section 3.1.1, we individually map each to a keyword encoding, as shown in Figure 3.2.1 below. \\

\begin{figure}
\center\includegraphics[width=1.0\textwidth]{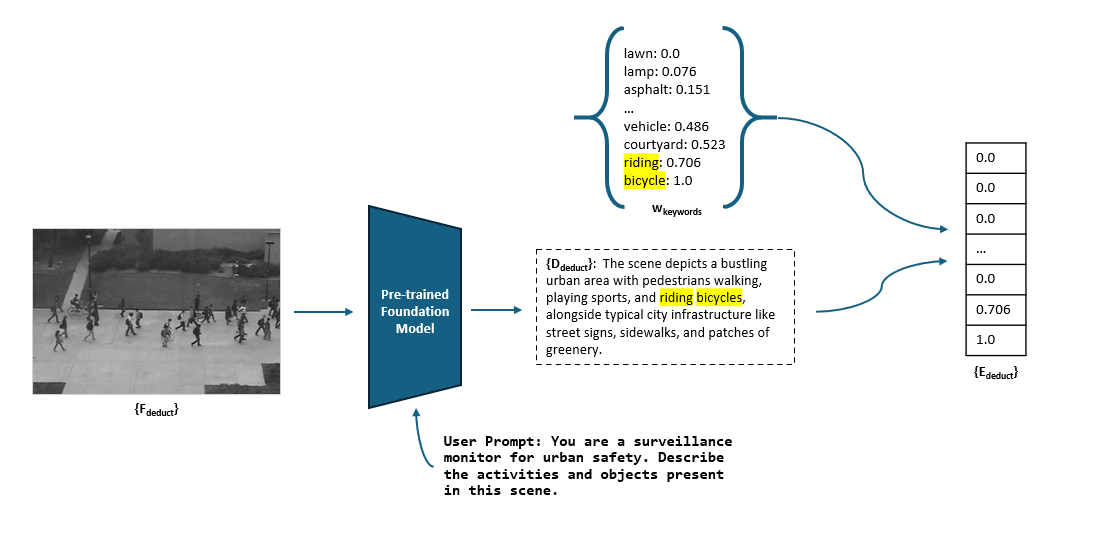}
\caption[Creating the keyword encoding. Description $D_{deduct}$ is generated by passing frame $F_{deduct}$ into the pre-trained foundational model. Then the description is mapped into a keyword encoding $E_{deduct}$ using the keyword weights $w_{keywords}$ from the induction stage.]{Creating the keyword encoding. Description $D_{deduct}$ is generated by passing frame $F_{deduct}$ into the pre-trained foundational model. Then the description is mapped into a keyword encoding $E_{deduct}$ using the keyword weights $w_{keywords}$ from the induction stage.}
\label{fig:create_keyword_encoding}
\end{figure}

We assign a weight for each keyword in the description to an encoding. This weighting keyword vector $w_{keywords}$ is pre-generated in induction. The resulting encoding $E_{deduct}$ reflects how strongly the generated frame description is associated with anomalies. The length of this encoding vector is equivalent to the number of elements in $w_{keywords}$. If a keyword is absent in the frame description, the respective component of the encoding is set to zero. This encoding is still interpretable to the user since we know each keyword's position in the encoding and weight value. Therefore, each encoding can be interpreted as the potential abnormality of the frame based on the presence of anomaly-related keywords.

\subsubsection{Binary Classification Model}
\begin{figure}
\center\includegraphics[width=1.0\textwidth]{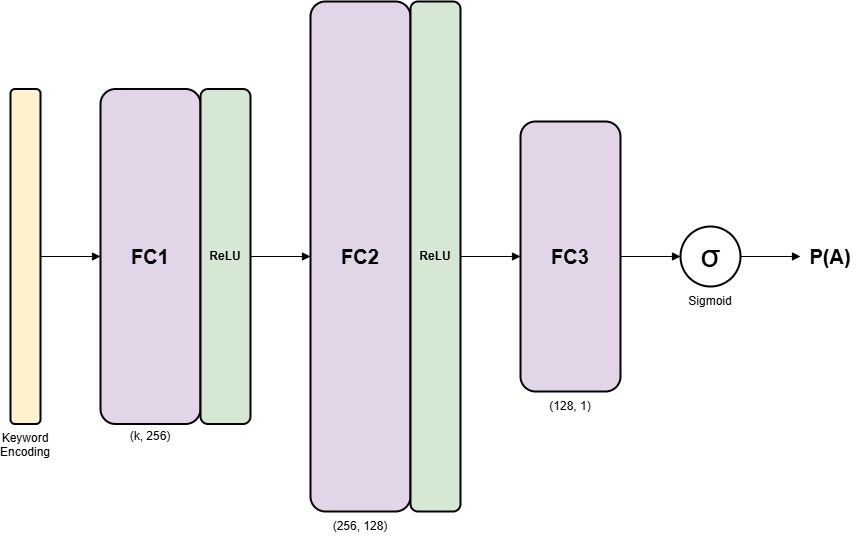}
\caption[Feed-forward network for binary classification. $FC-n$ stands for the fully-connected layer number $n$. The input and output dimensions are represented by $(i, o)$, with $i$ for input and $o$ for output. The dimension size $k$ is based on the number of keywords generated from induction. The probability $P(A)$ represents the chance that the inputted keyword encoding is an anomaly.]{Feed-forward network for binary classification. $FC-n$ stands for the fully-connected layer number $n$. The input and output dimensions are represented by $(i, o)$, with $i$ for input and $o$ for output. The dimension size $k$ is based on the number of keywords generated from induction. The probability $P(A)$ represents the chance that the inputted keyword encoding is an anomaly.}
\label{fig:feed_forward_network}
\end{figure}

The keyword encoding $E_{deduct}$ is fed into our binary classification model, designed to predict whether or not the frame input contains an anomaly. We decided to utilize a simple feed-forward neural network to satisfy this simple classification task. Our network includes three fully connected layers, as shown in Figure~\ref{fig:feed_forward_network}. The input dimension has $k$ neurons, where $k$ is the number of elements in the encoding, identical to the number of keywords identified in induction. The output layer has a single neuron that produces the anomaly probability of the frame input. \\

We train the model using this same pipeline. During this training process, our network learns the keyword encodings to output the final anomaly prediction and effectively learns the patterns and relationships in the data that indicate anomalies. Once trained, our model can take any keyword encoding and predict the likelihood that the frame is an anomaly. We then set a threshold the output must exceed to indicate an anomaly during test time.

\section{Experiments and Results}

This section will review the datasets selected for benchmarking our proposed method for video anomaly detection. Then, we will review the setup process of our experiments and the associated design choices. Next, we discuss the evaluation metrics necessary for measuring the success of our method. Classification results are reported and compared to other video anomaly detection methods. Finally, we compare qualitative features between different approaches and demonstrate the interpretability of our process with an example.

\subsection{Video Anomaly Detection Datasets}
In VAD, datasets are typically created from videos split into sequences of frames. These frames typically blend normal activities and abnormal activities or unusual events \citep{yossef2023review}. Most frames commonly include only normal activities since it is difficult to capture many anomalies due to the infrequency of their occurrence. Regardless, these datasets provide an excellent resource for models to differentiate routine occurrences from abnormal events. The three datasets we use, UCSD Ped2, ShanghaiTech, and CUHK Avenue \citep{mahadevan2010anomaly, liu2018future, lu2013abnormal}, are commonly used to develop robust video anomaly detection. Examples are included below in Figure~\ref{fig:benchmark_examples}.\\

\begin{figure}
\center\includegraphics[width=1.0\textwidth]{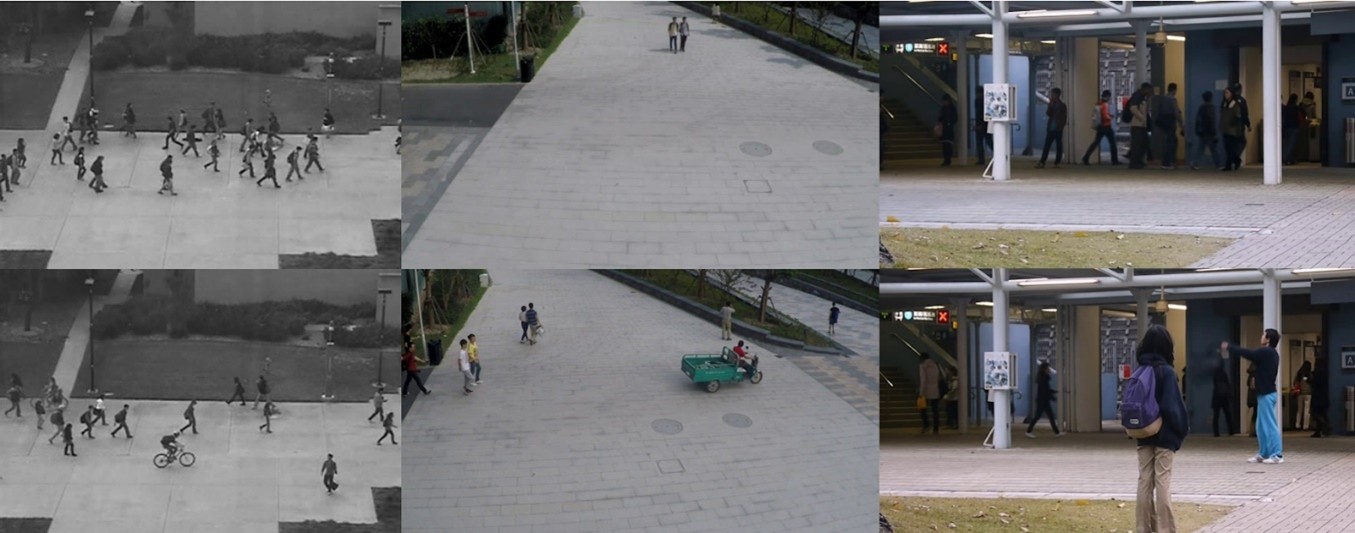}
\caption[VAD benchmark examples. These include UCSD Ped2 \citep{mahadevan2010anomaly} (left), ShanghaiTech \citep{liu2018future} (center), and the CUHK Avenue \citep{lu2013abnormal} (right) datasets. Images in the top row depict normal occurrences, and the bottom row depicts anomaly occurrences.]{VAD benchmark examples. These include UCSD Ped2 \citep{mahadevan2010anomaly} (left), ShanghaiTech \citep{liu2018future} (center), and the CUHK Avenue \citep{lu2013abnormal} (right) datasets. Images in the top row depict normal occurrences, and the bottom row depicts anomaly occurrences.}
\label{fig:benchmark_examples}
\end{figure}

The UCSD Ped2 dataset was gathered from a stationary camera mounted overlooking a pedestrian walkway at the University of California - San Diego campus. The dataset contains 16 videos for training and 12 videos for testing, including 12 abnormal events \citep{mahadevan2010anomaly}. The video should contain only pedestrians in the normal setting. Therefore, abnormal events like bikers, skaters, or cars can be described as any non-pedestrian entity on the walkway. Frame-level and pixel-level annotations are included, but we only utilize the frame-level annotations. UCSD Ped2 is a baseline for video anomaly detection because of its simplistic representation of an environment that lacks complex anomalies. However, many surveillance applications have a narrow scope like the one portrayed in this dataset, making it a suitable choice for testing detection effectiveness.\\

An alternative for video anomaly detection is the CUHK Avenue dataset, capturing scenes from an avenue at the Chinese University of Hong Kong. This dataset includes 16 training and 21 testing videos with 47 abnormal events, including loitering, throwing objects, and running \citep{liu2018future}. The advantage of this dataset comes from the increased complexity compared to UCSD Ped2 while maintaining well-defined individual anomalies. It is also worth noting that it includes additional complexity from camera shake and infrequent normal behavior in the dataset, which is important for measuring model robustness.\\

Created from scenes at the Shanghai Tech (SHTech) University Campus, the SHTech dataset provides various anomaly scenarios. Unlike the previous two datasets, this SHTech contains 13 scenes of multiple lighting conditions and camera angles \citep{lu2013abnormal}. There are 130 abnormal events with 274,515 training frames and 42,883 testing frames, making this significantly more significant than most video anomaly detection datasets. This allows us to test the adaptability of anomaly detection methods against a wide range of possible anomalies while providing more realistic scenarios compared to UCSD Ped2 and CUHK Avenue.

\subsection{Experimental Setup}
Using the Llama-3.2 Vision-Instruct-11B \cite{dubey2024llama} foundational model, we employ quantization for frame description generation to reduce computational and memory requirements. We convert the model's 16-bit float precision to 4-bit integer precision using Huggingface's BitsAndBytesConfig function \cite{huggingface}. This allows us to achieve faster inference for near-real-time prediction on resource-constrained devices when we deploy our system. \\

To obtain keywords that generalize the normal and anomalous behavior of the datasets, we determined that randomly sampling 20 videos each for both normal and anomalous frames performed well. This is because enough frames were sampled to learn typical behaviors from all the dataset's videos without over-fitting to any one specific occurrence in the videos. We omit these 40 randomly selected frames during classification model training/testing to maintain data integrity. \\

We employ the Scikit-Learn \cite{scikit-learn} library's TFidfVectorizer function for converting the corpus into a TF-IDF score matrix. We utilize function arguments to simplify extracting meaningful keywords from both corpora. The first one we set is 'stop\_words'. In NLP, words such as "the", "and", "is", and "or" are considered to be insignificant to the meaning of the sentence. Therefore, when we flag this argument, we omit such words from consideration for anomaly keywords, allowing our keyword weights to focus on critical parts of the frame description. Next, the TFidfVectorizer will enable us to set a ngram\_range from one to three. N-grams are collections of \textit{n} successive pieces of text. In our implementation, we maintain the default value of one to consider each keyword independently. This also increases the keyword encoding speed in deduction since we don't have to search the text for sets of words. An argument called max\_features is used to limit the number of words generated in the TF-IDF score matrix and, in our case, the number of keywords in our encodings. We limit this amount to 100 for the more straightforward UCSD Ped2 dataset and 200 for the more complex CUHK Avenue and SHTech datasets. The final arguments we employ are the min\_df and max\_df values. Since the TF-IDF score is between 0 and 1, we can adjust these limits to only output values in the matrix between these min\_df and max\_df values. In our implementation, we set the max\_df value on the UCSD Ped2 dataset to 0.95 for better performance and maintain the 0 to 1 range for the CUHK Avenue and UCSD Ped2 datasets.\\

Next, we made some design decisions for training and testing the binary classification model. We use a weighted Binary Cross-Entropy loss to address the common class imbalance issue in these video anomaly detection datasets. The positive class weight for each dataset was calculated as the inverse proportion of anomalous samples within the training set to provide further weight to the uncommon anomaly class during training. The model was initialized with the AdamW optimizer using a learning rate and decay rate of 0.001. To decrease the chance of overfitting, we employ a 5-fold cross-validation. Each fold was trained for a maximum of 20 epochs, and early stopping was used if the validation loss did not decrease for three consecutive epochs. We used custom batch sizes specific to the datasets, with 200 for UCSD Ped2, 1000 for CUHK Avenue, and 2000 frames for SHTech. The model with the best performance across the validation folds was selected for evaluation. We used 80\% of the frames for training and 20\% for testing.

\subsection{Evaluation Metrics}
For most implementations of video anomaly detection, the frame-level area under the receiver operation characteristic (AUROC) is used to evaluate the ground truth labels. AUROC measures how well a classification model can distinguish the positive and negative class in binary classification by plotting the True Positive Rate (TPR) against the False Positive Rate (FPR). VAD employs this metric over accuracy because of its improved reliability when dealing with imbalanced datasets, where one class occurs more often than the other, which is common in VAD datasets. AUROC typically comes in two forms: micro-averaged, where the score is computed on all frames from all the videos, whereas the macro-averaged form computes the score separately for each video and then takes the average of them \citep{doshi2023towards}.

\subsection{Ablation Study}

We evaluate the impact of the foundational model choice on anomaly detection performance, by conducting an ablation study comparing the Llama-3.2 Vision-Instruct-11B \cite{dubey2024llama} and OpenBmb Mini-CPM \cite{hu2024minicpm} models. Table ~\ref{tab:ablation} depicts our results on the UCSD Ped2, ShanghaiTech, and CUHK Avenue datasets.

\begin{table}[ht]
\centering
\caption[Ablation study results. Each cell shows the AUROC (\%) / inference speed in seconds per frame for different foundational models and datasets.]{Ablation study results. Each cell shows the AUROC (\%) / inference speed in seconds per frame for different foundational models and datasets.}
\begin{tabular}{|l|c|c|c|}
\hline
& \textbf{UCSD Ped2 \citep{mahadevan2010anomaly}} & \textbf{ShanghaiTech \citep{liu2018future}} & \textbf{CUHK Avenue \citep{lu2013abnormal}}\\ \hline
Vision-Instruct-11B \citep{dubey2024llama} & 0.865/5.77s & 0.753/5.17s & 0.742/5.38s\\
Mini-CPM \citep{hu2024minicpm} & 0.865/2.43s & 0.707/2.09s & 0.604/2.12s \\
\hline
\end{tabular}
\label{tab:ablation}
\end{table}

These results show a trade-off between detection accuracy and speed. Both models achieve comparable performance on the UCSD Ped2 dataset, but Vision-Instruct-11B performs noticeably better on the more complex ShanghaiTech and CUHK Avenue datasets. Specifically, Vision-Instruct-11B achieves ROC-AUC scores of 74.2\% and 75.3\% on ShanghaiTech and CUHK Avenue compared to Mini-CPM's 60.4\% and 70.7\%. This can be explained by the increase in detail in Vision-Instruct's frame descriptions which allow it to capture more subtleties compared to Mini-CPM. The improved accuracy and detail come at the cost of computational overhead and speed due to Vision-Instruct's longer generation times. Mini-CPM exhibits significantly faster inference across all of the datasets, slightly above two seconds per frame compared to Vision-Instruct's five or six seconds per frame. Mini-CPM also defaults to using 4-bit int precision and fewer parameters than Vision-Instruct-11B, making it more viable on constrained systems. \\

Therefore, the choice of a foundational model depends on the specific requirements of the application. If high accuracy is critical, even at the expense of computational speed, Vision-Instruct-11B is the better option. Likewise, if real-time performance or resource constraints are critical, Mini-CPM offers an alternative with a reduced computational footprint at the cost of a potential accuracy decrease on complex scenarios.

\subsection{Classification Results}

As seen in Table ~\ref{tab:quantitative}, we compare the frame-level AUROC between our method and multiple state-of-the-art approaches across the three selected benchmark datasets: UCSD Ped2, ShanghaiTech, and CUHK Avenue \citep{mahadevan2010anomaly, liu2018future, lu2013abnormal}. Our results demonstrate that while Video Anomaly Detection with Structured Keywords (VADSK) doesn't outperform most advanced SOTA methods, it achieves competitive performance in certain scenarios, such as on the ShanghaiTech dataset, where it achieves an AUROC of 75.3\%. This is comparable to methods such as HF2-VAD with 76. 2\% and exceeds Toward Interpretable VAD (68. 9\%). It is important to understand that our results on the UCSD Ped2 and CUHK Avenue datasets still underperform the top-performing methods, which achieve scores above 90\%. \\

Our performance disparity between UCSD Ped2 and the other two benchmarks can be explained by their increased anomaly event complexity, indicating that our method is better suited for handling simpler anomaly events and that there is room for improvement in our method's ability to generalize across many diverse anomaly contexts. \\

\begin{table}[ht]
\centering
\caption{Frame-level AUROC (\%) Comparison}
\begin{tabular}{|l|c|c|c|}
\hline
 & \textbf{UCSD Ped2 \citep{mahadevan2010anomaly}} & \textbf{ShanghaiTech \citep{liu2018future}} & \textbf{CUHK Avenue \citep{lu2013abnormal}}\\ \hline
HF2-VAD \citep{liu2021hybrid} & 0.993 & 0.762 & 0.911\\
Towards Interpretable VAD \citep{doshi2023towards} & - & 0.689 & 0.790 \\
TEVAD \citep{chen2023tevad} & 0.987 & \textbf{0.981} & - \\
Attribute-based VAD \citep{reiss2022attribute} & 0.991 & 0.859 & \textbf{0.937} \\
MULDE \citep{micorek2024mulde} & \textbf{0.997} & 0.864 & 0.931 \\
AnomalyRuler \citep{yang2025follow} & 0.965 & 0.852 & 0.822 \\ \hline
VADSK (ours) & 0.865 & 0.753 & 0.742 \\ 
\hline
\end{tabular}
\label{tab:quantitative}
\end{table}

\subsection{Qualitative Comparison and Analysis}

While the quantitative performance of our approach doesn't outperform state-of-the-art methods across the datasets, as seen in Table ~\ref{tab:quantitative}, the significance of our approach is how we developed a memory-efficient, interpretable system with real-time inference capability. The methods we compare against ours have some, but not all, of these necessary traits, as depicted in Table ~\ref{tab:qualitative}.

\begin{table}[ht]
\centering
\caption{Qualitative Comparison}
\begin{tabular}{|l|c|c|c|}
\hline
\textbf{} & \textbf{Interpretable} & \textbf{Real-time} & \textbf{Memory-efficient}\\ \hline
HF2-VAD \citep{liu2021hybrid} & \xmark & \xmark & \xmark\\
Towards Interpretable VAD \citep{doshi2023towards} & \cmark & \xmark & \cmark\\
TEVAD \citep{chen2023tevad} & \cmark &\xmark & \xmark\\
Attribute-based VAD \citep{reiss2022attribute} & \cmark & \xmark & \xmark\\
MULDE \citep{micorek2024mulde} & \xmark & \cmark & \cmark\\
AnomalyRuler \citep{yang2025follow} & \cmark & \xmark & \xmark\\ \hline
VADSK (ours) &  \cmark &  \cmark &  \cmark\\
\hline
\end{tabular}
\label{tab:qualitative}
\end{table}

\begin{figure}
\center\includegraphics[width=1.0\textwidth]{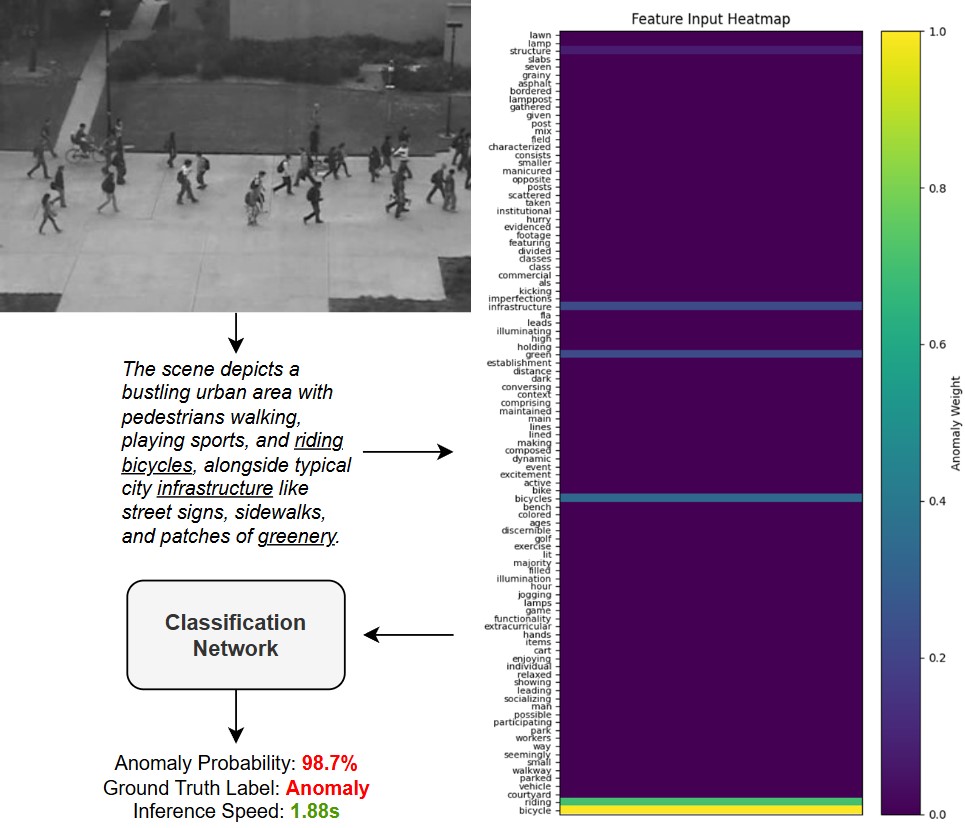}
\caption[Interpretable Inference. The heatmap (right) visualizes a keyword encoding for classification derived from the frame description (center-left). The frame description was generated by passing the frame (top-left) into the MiniCPM foundational model \citep{hu2024minicpm}. The different colors in this heatmap represent different weight values between 0 and 1. Finally, the result is a probability (bottom-left) that an anomaly has occurred.]{Interpretable Inference. The heatmap (right) visualizes a keyword encoding for classification derived from the frame description (center-left). The frame description was generated by passing the frame (top-left) into the MiniCPM foundational model \citep{hu2024minicpm}. The different colors in this heatmap represent different weight values between 0 and 1. Finally, the result is a probability (bottom-left) that an anomaly has occurred.}
\label{fig:interpretable_inference}
\end{figure}

Interpretability is an important aspect of our system, defined by the user's ability to interpret the extracted features and understand how they are used in the classification decision-making process. Typically, this is done with textual information incorporated into the pipeline \citep{yang2025follow, doshi2023towards, chen2023tevad} or with visual information such as velocity and pose \citep{reiss2022attribute}. Our system is interpretable since we use frame descriptions and encode them based on pre-defined keyword weights that are transparent to the user. As seen in Figure ~\ref{fig:interpretable_inference}, it is possible for the user to view the steps used in outputting the final prediction and adjust the keyword encoding or foundational model for frame description generation for improved results. Such transparency is vital in scenarios where understanding the reason behind an anomaly is equally important as detection. \\

Another advantage our system provides is the near real-time inference. We define real-time systems as the ability to run inference with minimal latency on individual or windows of frames \citep{micorek2024mulde}, compared to inputting the entire video at once to generate a prediction. Our approach is real-time because we generate descriptions for each frame, pass in the respective keyword encoding one at a time during inference, and receive predictions at most a few seconds after input. This proves valuable in applications that need immediate or rapid responses to detected anomalies. \\

Lastly, our system maintains memory efficiency during the induction and deduction stages. We define memory-efficient systems as those that do not require the storage of temporal information and do not run expensive feature extraction processes in parallel with one another \citep{micorek2024mulde, doshi2023towards}. Our approach is memory-efficient since we do not require temporal information and use one sequential process for feature extraction and classification. Most of our computational overhead comes from the quantized foundation model, which generates initial frame descriptions from the inputted frame. This efficiency is necessary for many video anomaly detection systems since they are deployed in resource-constrained environments and must be scaled for large-scale surveillance systems. \\

Combining these three traits - interpretability, near real-time inference, and efficiency makes VADSK a practical solution for video anomaly use cases that require these critical capabilities at the cost of decreased performance against SOTA methods. Our work, therefore, represents an essential contribution by offering a more balanced approach that addresses these key considerations for the real-world application of video anomaly detection systems.

\section{Conclusion}
This paper presents a novel approach to video anomaly detection using structured keywords, demonstrating the potential for exclusively using text-based features for detection. We developed a lightweight interpretable pipeline for video anomaly detection consisting of an induction and deduction stage. Our method achieves comparable performance to state-of-the-art methods on certain benchmarks and demonstrates the feasibility of real-time inference without temporal information. Our performance gap on more complex datasets, such as CUHK Avenue and ShanghaiTech, shows that there is still more room for improvement. Our work has broader implications. The increased interpretability could improve trust and adoption of video surveillance systems for public spaces and critical infrastructure. Our simple, lightweight pipeline can make such systems suitable for edge devices with limited computational resources or large-scale networks, enabling adoption in environments previously limited by hardware constraints or scalability issues. Finally, our near real-time inference allows for immediate response times to anomalous events, which can be critical in detecting security threats or other emergencies. \\

For future work, improving the keyword generation and selection process would be worthwhile. The selection process could include advanced natural language processing techniques for selecting nuanced keywords, dynamically updating the keyword selection from patterns emerging in the video footage, or generating domain-specific keywords depending on the particular environment. Experimenting with different foundational models and classification architectures would help improve the frame description quality and discriminative capabilities. Finally, it could be beneficial to test the effectiveness of our method on specialized domains such as industrial safety, healthcare, or traffic to investigate the effectiveness of our method in real-world scenarios. By leveraging the power of foundational models and natural language processing, we have opened up new possibilities for video anomaly detection to be deployed in real-world scenarios, paving the way for intelligent and responsive surveillance for improving urban safety.

\bibliographystyle{plain}
\bibliography{references}

\begin{thebibliography}{19}
\providecommand{\natexlab}[1]{#1}
\providecommand{\url}[1]{\texttt{#1}}
\expandafter\ifx\csname urlstyle\endcsname\relax
  \providecommand{\doi}[1]{doi: #1}\else
  \providecommand{\doi}{doi: \begingroup \urlstyle{rm}\Url}\fi

\bibitem[Chen et~al.(2023)Chen, Ma, Yew, Hur, and Khoo]{chen2023tevad}
W.~Chen, K.~T. Ma, Z.~J. Yew, M.~Hur, and D.~A. A. Khoo.
\newblock Tevad: Improved video anomaly detection with captions.
\newblock In \emph{Proceedings of the IEEE/CVF Conference on Computer Vision
  and Pattern Recognition}, pages 5549--5559, 2023.

\bibitem[Doshi and Yilmaz(2023)]{doshi2023towards}
K.~Doshi and Y.~Yilmaz.
\newblock Towards interpretable video anomaly detection.
\newblock In \emph{Proceedings of the IEEE/CVF Winter Conference on
  Applications of Computer Vision}, pages 2655--2664, 2023.

\bibitem[Dubey et~al.(2024)Dubey, Jauhri, Pandey, Kadian, Al-Dahle, Letman,
  and others]{dubey2024llama}
A.~Dubey, A.~Jauhri, A.~Pandey, A.~Kadian, A.~Al-Dahle, A.~Letman, and others.
\newblock The llama 3 herd of models.
\newblock \emph{arXiv preprint arXiv:2407.21783}, 2024.

\bibitem[Hu et~al.(2024)Hu, Tu, Han, He, Cui, Long, and others]{hu2024minicpm}
S.~Hu, Y.~Tu, X.~Han, C.~He, G.~Cui, X.~Long, and others.
\newblock Minicpm: Unveiling the potential of small language models with
  scalable training strategies.
\newblock \emph{arXiv preprint arXiv:2404.06395}, 2024.

\bibitem[Karim et~al.(2024)Karim, Doshi, and Yilmaz]{karim2024real}
H.~Karim, K.~Doshi, and Y.~Yilmaz.
\newblock Real-time weakly supervised video anomaly detection.
\newblock In \emph{Proceedings of the IEEE/CVF winter conference on
  applications of computer vision}, pages 6848--6856, 2024.

\bibitem[Liu et~al.(2018)Liu, Luo, Lian, and Gao]{liu2018future}
W.~Liu, W.~Luo, D.~Lian, and S.~Gao.
\newblock Future frame prediction for anomaly detection--a new baseline.
\newblock In \emph{Proceedings of the IEEE conference on computer vision and
  pattern recognition}, pages 6536--6545, 2018.

\bibitem[Liu et~al.(2021)Liu, Nie, Long, Zhang, and Li]{liu2021hybrid}
Z.~Liu, Y.~Nie, C.~Long, Q.~Zhang, and G.~Li.
\newblock A hybrid video anomaly detection framework via memory-augmented flow
  reconstruction and flow-guided frame prediction.
\newblock In \emph{Proceedings of the IEEE/CVF international conference on
  computer vision}, pages 13588--13597, 2021.

\bibitem[Lu et~al.(2013)Lu, Shi, and Jia]{lu2013abnormal}
C.~Lu, J.~Shi, and J.~Jia.
\newblock Abnormal event detection at 150 fps in matlab.
\newblock In \emph{Proceedings of the IEEE international conference on computer
  vision}, pages 2720--2727, 2013.

\bibitem[Mahadevan et~al.(2010)Mahadevan, Li, Bhalodia, and
  Vasconcelos]{mahadevan2010anomaly}
V.~Mahadevan, W.~Li, V.~Bhalodia, and N.~Vasconcelos.
\newblock Anomaly detection in crowded scenes.
\newblock In \emph{2010 IEEE Computer Society Conference on Computer Vision
  and Pattern Recognition}, pages 1975--1981, 2010.

\bibitem[Micorek et~al.(2024)Micorek, Possegger, Narnhofer, Bischof, and
  Kozinski]{micorek2024mulde}
J.~Micorek, H.~Possegger, D.~Narnhofer, H.~Bischof, and M.~Kozinski.
\newblock Mulde: Multiscale log-density estimation via denoising score matching
  for video anomaly detection.
\newblock In \emph{Proceedings of the IEEE/CVF Conference on Computer Vision
  and Pattern Recognition}, pages 18868--18877, 2024.

\bibitem[Nam et~al.(2024)Nam, Yoo, and Hong]{nam2024log}
S.~Nam, J.~H. Yoo, and J.~W. K. Hong.
\newblock Log-tf-idf for anomaly detection in network switches.
\newblock In \emph{NOMS 2024-2024 IEEE Network Operations and Management
  Symposium}, pages 1--9, 2024.

\bibitem[Pedregosa et~al.(2011)Pedregosa, Varoquaux, Gramfort, Michel,
  Thirion, Grisel, Blondel, Prettenhofer, Weiss, Dubourg, Vanderplas, Passos,
  Cournapeau, Brucher, Perrot, and Duchesnay]{scikit-learn}
F.~Pedregosa, G.~Varoquaux, A.~Gramfort, V.~Michel, B.~Thirion, G.~Grisel,
  M.~Blondel, P.~Prettenhofer, R.~Weiss, V.~Dubourg, J.~Vanderplas, A.~Passos,
  D.~Cournapeau, M.~Brucher, M.~Perrot, and E.~Duchesnay.
\newblock Scikit-learn: Machine learning in {P}ython.
\newblock \emph{Journal of Machine Learning Research}, 12:\penalty0 2825--2830,
  2011.

\bibitem[Reiss and Hoshen(2022)]{reiss2022attribute}
T.~Reiss and Y.~Hoshen.
\newblock Attribute-based representations for accurate and interpretable video
  anomaly detection.
\newblock \emph{arXiv preprint arXiv:2212.00789}, 2022.

\bibitem[Sabokrou et~al.(2018)Sabokrou, Khalooei, Fathy, and
  Adeli]{sabokrou2018adversarially}
M.~Sabokrou, M.~Khalooei, M.~Fathy, and E.~Adeli.
\newblock Adversarially learned one-class classifier for novelty detection.
\newblock In \emph{Proceedings of the IEEE conference on computer vision and
  pattern recognition}, pages 3379--3388, 2018.

\bibitem[Sandhu and Mohammed(2022)]{sandhu2022detecting}
A.~Sandhu and S.~Mohammed.
\newblock Detecting anomalies in logs by combining nlp features with embedding
  or tf-idf, 2022.

\bibitem[Sultani et~al.(2018)Sultani, Chen, and Shah]{sultani2018real}
W.~Sultani, C.~Chen, and M.~Shah.
\newblock Real-world anomaly detection in surveillance videos.
\newblock In \emph{Proceedings of the IEEE conference on computer vision and
  pattern recognition}, pages 6479--6488, 2018.

\bibitem[Wu et~al.(2020)Wu, Liu, Shi, Sun, Shao, Wu, and Yang]{wu2020not}
P.~Wu, J.~Liu, Y.~Shi, Y.~Sun, F.~Shao, Z.~Wu, and Z.~Yang.
\newblock Not only look, but also listen: Learning multimodal violence
  detection under weak supervision.
\newblock In \emph{Computer Vision--ECCV 2020: 16th European Conference,
  Glasgow, UK, August 23--28, 2020, Proceedings, Part XXX 16}, pages
  322--339. Springer International Publishing, 2020.

\bibitem[Wu et~al.(2024)Wu and others]{wu2024deep}
P.~Wu and others.
\newblock Deep learning for video anomaly detection: A review.
\newblock \emph{arXiv preprint arXiv:2409.05383}, 2024.

\bibitem[Yang et~al.(2025)Yang, Lee, Dariush, Cao, and Lo]{yang2025follow}
Y.~Yang, K.~Lee, B.~Dariush, Y.~Cao, and S.~Y. Lo.
\newblock Follow the rules: reasoning for video anomaly detection with large
  language models.
\newblock In \emph{European Conference on Computer Vision}, pages 304--322,
  2025.

\bibitem[Yossef et~al.(2023)Yossef, Gamal, Abdel-Kader, and Ali]{yossef2023review}
I.~M. Yossef, M.~Gamal, R.~F. Abdel-Kader, and K.~A. E. Ali.
\newblock A review on video anomaly detection datasets.
\newblock \emph{Suez Canal Engineering, Energy and Environmental Science},
  1\penalty0 (2):\penalty0 1--9, July 2023.

\end{thebibliography}
\end{document}